\documentclass[letterpaper, 10 pt, conference, twocolumn]{ieeeconf}

\IEEEoverridecommandlockouts
\usepackage{cite}
\usepackage{amsmath,amssymb,amsfonts}
\usepackage{multirow}
\usepackage{algorithm}
\usepackage{algorithmic}
\usepackage{graphicx}
\usepackage{textcomp}

\usepackage{multirow}
\usepackage[table,xcdraw]{xcolor}

\usepackage{soul}
\usepackage{subcaption}

\def\BibTeX{{\rm B\kern-.05em{\sc i\kern-.025em b}\kern-.08em T\kern-.1667em\lower.7ex\hbox{E}\kern-.125emX}} 

\definecolor{deep-red}{RGB}{192, 0, 0}
\definecolor{deep-purple}{RGB}{120, 0, 170}
\definecolor{good-green}{RGB}{0,175,0} 
\definecolor{purple}{RGB}{210, 0, 210}

\newcommand{\ch}[1]{\textcolor{black}{#1}}

\makeatletter
\let\NAT@parse\undefined
\makeatother
\usepackage[colorlinks]{hyperref}

\begin{document}
\setlength{\textfloatsep}{8pt plus 2pt minus 4pt}

\title{Improving the realism of robotic surgery simulation through injection of learning-based estimated errors}
\author{%
     Juan Antonio Barragan$^{1}$, Hisashi Ishida$^{1}$, Adnan Munawar$^1$, and Peter Kazanzides$^1$%
     \thanks{$^1$Department of Computer Science, Johns Hopkins University, Baltimore, MD 21218, USA.  
     Email: {\tt \{jbarrag3,hishida3,pkaz\}@jhu.edu}}
}

\maketitle

\begin{abstract}
The development of algorithms for automation of subtasks during robotic surgery can be accelerated by the availability of realistic simulation environments. In this work, we focus on one aspect of the realism of a surgical simulator, which is the positional accuracy of the robot. In current simulators, robots have perfect or near-perfect accuracy, which is not representative of their physical counterparts. We therefore propose a pair of neural networks, trained by data collected from a physical robot, to estimate both the controller error and the kinematic and non-kinematic error. These error estimates are then injected within the simulator to produce a simulated robot that has the characteristic performance of the physical robot. In this scenario, we believe it is sufficient for the estimated error used in the simulation to have a statistically similar distribution to the actual error of the physical robot. This is less stringent, and therefore more tenable, than the requirement for error compensation of a physical robot, where the estimated error should equal the actual error.
Our results demonstrate that error injection reduces the mean position and orientation differences between the simulated and physical robots from 5.0\,mm / 3.6\,deg to 1.3\,mm / 1.7\,deg, respectively, which represents reductions by factors of 3.8 and 2.1.

\end{abstract}

\section{Introduction}
In robotic minimally invasive surgery (RMIS), automation of surgical subtasks holds the potential to improve the surgical workflow and reduce surgeon's workload \cite{attanasioAutonomySurgicalRobotics2021}. Surgical robotic systems dealing with rigid anatomical structures, such as orthopedic systems, have already achieved high levels of autonomy \cite{yip_robot_2017}. However, it remains an open challenge to reach these same levels of autonomy in soft tissue surgery due to perception and modeling challenges.

Technical challenges of autonomy in soft-tissue surgeries have led manufacturers of surgical robotic platforms to focus on developing systems that surgeons can fully teleoperate to perform the procedures. In this regard, these surgical robotics systems are not designed to have perfect absolute positioning accuracy, but rather an intuitive teleoperation interface that allows the surgeon to visually close the loop to perform actions in the surgical workspace. One of the most successful systems following this design paradigm is the da Vinci surgical system (Intuitive Surgical, Inc., Sunnyvale, CA, USA). 

The limited positioning accuracy of surgical robots presents additional challenges for developing autonomous algorithms in surgery and has been mainly approached with two different strategies.
The first strategy has been to develop robust calibration procedures to ensure that the surgical robot will reach a commanded pose accurately. These calibration procedures often require a learning-based model that can learn to correct for the kinematic and non-kinematic errors of the robot \cite{peng_real-time_2020, hwang_efficiently_2020, seita_fast_2018}. The second strategy has been to develop sensor-based closed-loop control algorithms that ensure that the surgical robot will reach the desired target pose despite being inaccurate. Techniques such as visual-servoing and reinforcement learning would fall under this category \cite{paradis_intermittent_2021, li_super_2020}. Given the inherently limited absolute positioning accuracy of teleoperated surgical robots and the difficulties of developing calibration procedures that will compensate for errors in all situations, we argue that favoring algorithms that can tolerate realistic amounts of robot error is a more tenable solution in the long-term to introduce autonomy in surgery. 

Simulators have historically been accelerators for the development of algorithms in robotics, however, they are currently not well suited to encourage the development of autonomous algorithms that can work with inaccurate robots. This is mainly the case because robotic simulators use an ideal mathematical model of the robot that does not account for real sources of errors such as manufacturing tolerances, wear and tear, and mechanical compliance of the system \cite{chrysilla_compliance_2019,peng_real-time_2020,hwang_efficiently_2020}. These deviations between the real and the simulated system imply that algorithms developed on the simulation will have a reduced performance when deployed on the inaccurate real system \cite{hofersim2real2021}. In this work, we propose to address this limitation by taking a \textit{real2sim} approach where we model the positioning error in a real robot and use this model to inject realistic errors into a simulated robot. 

This approach may raise the obvious question of why we are not using our error models to correct the error in the real robot so that autonomous algorithms developed with an accurate simulated robot can then be deployed on an accurate real robot. The answer again is that it is easier (i.e., more reliable) to break an accurate robot than it is to fix an inaccurate robot.  Specifically, error compensation requires that the error estimate be accurate all the time, whereas we contend that error injection (in particular, for training machine learning algorithms) requires only that the error estimate be statistically similar to the actual error.

 \begin{figure*}[ht]
      \centering
      \includegraphics[width=1.0\textwidth]{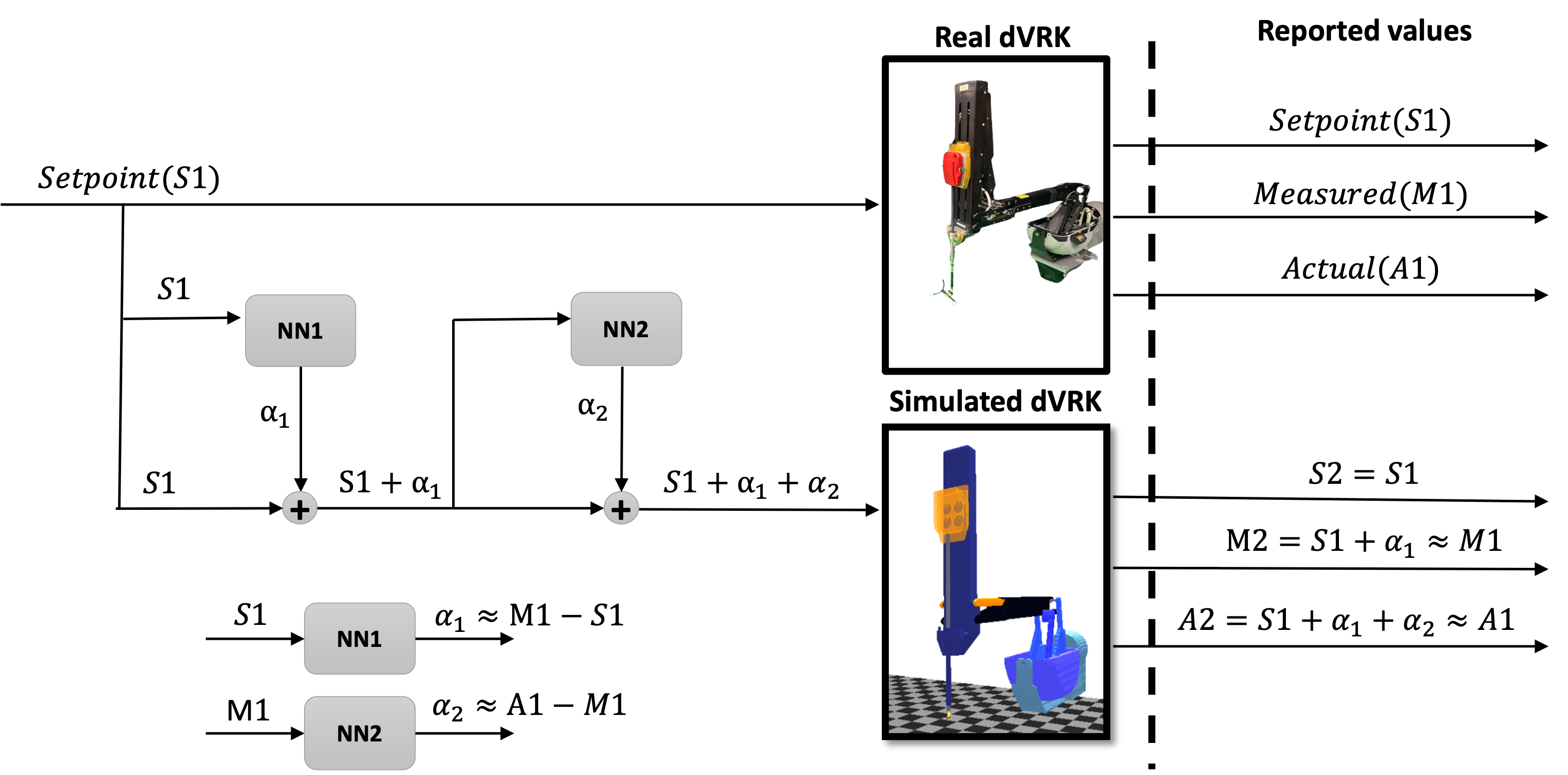}
      \captionof{figure}{Proposed system to emulate the real-robot motion patterns using a simulated dVRK.}
      \label{fig:system_overview}
 \end{figure*}
 
With these considerations in mind, we propose an error injection pipeline that can improve the realism of robotic surgery simulation with the goal of encouraging researchers to develop novel algorithms that can tolerate realistically inaccurate simulated robots. As an approach to model the robot's positioning error, we propose to split errors into errors due to the robot's control system and errors due to kinematic and non-kinematic imprecision. This is a reasonable distinction as the former errors are attributed to the control software running on the robot, while the latter errors are related to mechanical defects and manufacturing tolerances that are specific to each robot. After modeling each source of error independently with a multi-layer perceptron (MLP), errors can then be injected into the simulated robot to obtain realistic motion patterns. This proposed error injection pipeline was tested on a real and simulated da Vinci Research
Kit (dVRK) \cite{kazanzides_open-source_2014}. \ch{Additionally, the code and datasets used in this work are provided as a public GitHub repository.} In summary, this paper presents the following contributions:
 
\begin{enumerate}
    \item A general methodology to model the kinematic error in surgical robots by splitting positioning error into controller-related error and error related to the system's mechanical limitations. 
    \item Trained neural networks to model and predict positioning error for a robot. 
    \item Implementation of an error injection algorithm that ensures that a simulated robot will have positioning error patterns that are similar to the real robot.
\end{enumerate}

\section{Related Work}

Several efforts have been made to model the nonlinear kinematic errors of surgical robots, which is relevant to our work even though those efforts focused on using those models to correct the error, rather than to inject it into a simulated robot.

One traditional approach has been to develop explicit models for specific sources of error. For instance,  Miyasaka \textit{et al.} explicitly modeled physical effects of cable-driven mechanisms with a hysteretic cable stretch model and a cable-pulley network friction model~\cite{Miyasaka2020Modeling}. Chrysilla \textit{et al.} proposed a compliance model that related joint motion with lateral forces on the shaft and corrected for non-kinematic errors due to external forces. 

Alternatively, robot errors can also be modeled using data-driven approaches such as Gaussian process \cite{Mahler2014Learning} regression or Deep Neural Networks \cite{Seita2017FastAR}, \cite{peng2023ablation}. 
In particular, Peng \textit{et al.} \cite{peng_real-time_2020} proposed using a multi-layer perceptron (MLP) to model the errors of a Raven robot\cite{hannaford_raven-ii_2013} in Cartesian space. Hwang \textit{et al.}\cite{hwang_efficiently_2020} advanced on this idea by training a recurrent neural network that included current and previous robot states to model robot errors in joint space. Following \cite{hwang_efficiently_2020}, we decided to perform modeling of the error via joint offsets; however, we decided to utilize a simpler MLP model as our work only focused on modeling the robot's static error.

\section{Methodology}
 
\subsection{Problem definition}

The main goal of this paper is to ensure that a simulated robot can replicate the position error of its real counterpart. To achieve this goal, we first make the distinction between three different end-effector poses that exist in a real robotic system. The first pose is referred to as the setpoint pose $(S_1)$, which is the target pose given to the robot's control system to initiate a motion. The next pose is the measured pose $(M_1)$. This is the pose that can be calculated with a measurement from the robot's encoders and the forward kinematic model. Lastly, we have the actual pose $(A_1)$, which is the pose to which the robot has actually moved. In general, this pose is not known, but it can be measured by observing the gripper with an external sensor, such as an optical tracker.

In an ideal system, $S_1$, $M_1$, and $A_1$ would be identical, but due to limitations of real mechanical and control systems, there will always be discrepancies between the three of them. Assuming the measured and actual poses of the simulated robot are $M_2$ and $A_2$, the goal of this paper is to ensure that the statistical distributions of $M_2^k-S_2^k \approx M_1^k-S_1^k$, and $A_2^k-M_2^k \approx A_1^k-M_1^k$, when the real and simulated robots are commanded to move through the same sequence of $N$ setpoints, $S_2^k=S_1^k$, $k = 1..N$.
A stricter, but not necessary, requirement would be for $M_2^k \approx M_1^k$ and $A_2^k \approx A_1^k$ for all values of $k$.
In this study, we consider only the static position accuracy of the robot (i.e., when it is no longer moving), primarily due to the difficulty of measuring the true pose of the real robot ($A_1$) during motion.

Given that the simulated robot does not suffer from positioning error, we relied on two neural networks to modify the setpoint provided to the simulated robot such that it matches the motion of the real robot. The first neural network is trained to predict the controller error (differences between $M_1$ and $S_1$), while the second network predicts differences between the actual and measured poses. \ch{Since both the real and simulated robots are controlled in joint space, the neural networks are also trained to predict joint offsets.} The complete correction framework can be seen in Figure \ref{fig:system_overview}. 

The rest of the methodology section is divided as follows. Subsection \ref{metho-A} describes how measured ($M_1$) and setpoint $(S_1)$ poses are collected to train the first neural network. Subsection \ref{metho-B} explains the procedure to register the robot and tracker and then how to use this information to calculate the dVRK grippers' actual pose ($A_1$). Subsections \ref{metho-C} and \ref{metho-D} explain how neural networks are trained to predict the robot's positioning error from the generated data and the evaluation metrics for each model. Lastly, subsection \ref{metho-E} describes how the neural network models are incorporated into the simulated robot. 

\subsection{Mathematical notation}
\ch{In this paper, poses are represented using a $4\times4$ rigid transformation matrix $T$ that contains a rotational component $R$ and a translational component $t$. To describe the relative pose of coordinate frame $i$ with respect to coordinate frame $j$, we use the notation ${}^{j}T_{i}$. Lastly, we denote the i$^{th}$ measured joint value which can be read from the robot's encoders as $q_i$. Since the dVRK manipulators have only 6 joints, we used the values $q_1$ to $q_6$ in this work.}

\subsection{Modeling errors between setpoint and measured pose }\label{metho-A}

In dVRK, controller errors, or errors between setpoint poses and measured poses, can be easily measured by collecting the last setpoint that was provided to the robot and measuring the position of the encoders. 
Historically, the dVRK had large differences between $S_1$ and $M_1$ because it used a proportional-derivative (PD) controller that depended on steady-state error to compensate for gravity. Recently, a disturbance observer was added to the dVRK PD controller \cite{Yilmaz2023}, and used for this work, which reduces but does not eliminate the error.


\subsection{Modeling errors between measured and actual pose }\label{metho-B}

In dVRK, the measured pose $(M_1)$ is calculated by using the measured joint angles from the encoders and the forward kinematics function, which is parameterized by the robot's Denavit–Hartenberg (DH) parameters. Differences between $A_1$ and $M_1$ can be broadly attributed to kinematic and non-kinematic errors. Kinematic errors exist due to deviations from the nominal robot kinematic parameters, which can occur due to manufacturing tolerances and wear and tear of the system. Non-kinematic errors are due to cable-related effects, e.g., hysteresis and cable tension, and mechanical deformations due to external forces applied to the instrument arm \cite{chrysilla_compliance_2019}.

In this paper, we aim to calculate an actual pose $(A_1)$ that incorporates corrections for both kinematic and non-kinematic errors. To calculate $A_1$, we attach an optically tracked marker to the robot's final link. Then, we perform a hand-eye calibration to identify the rigid transformation between the marker and the gripper, and the optical tracker and the robot. Lastly, $A_1$ is calculated using the hand-eye transformations and a measurement from the tracking system.  For this paper, data is only collected after the robot comes to a complete stop to avoid synchronization issues between the tracking system and the robot.

\subsubsection{Hand-eye calibration and actual pose estimation} 

\begin{figure}[ht]
    \centering
    \includegraphics[width=0.48\textwidth]{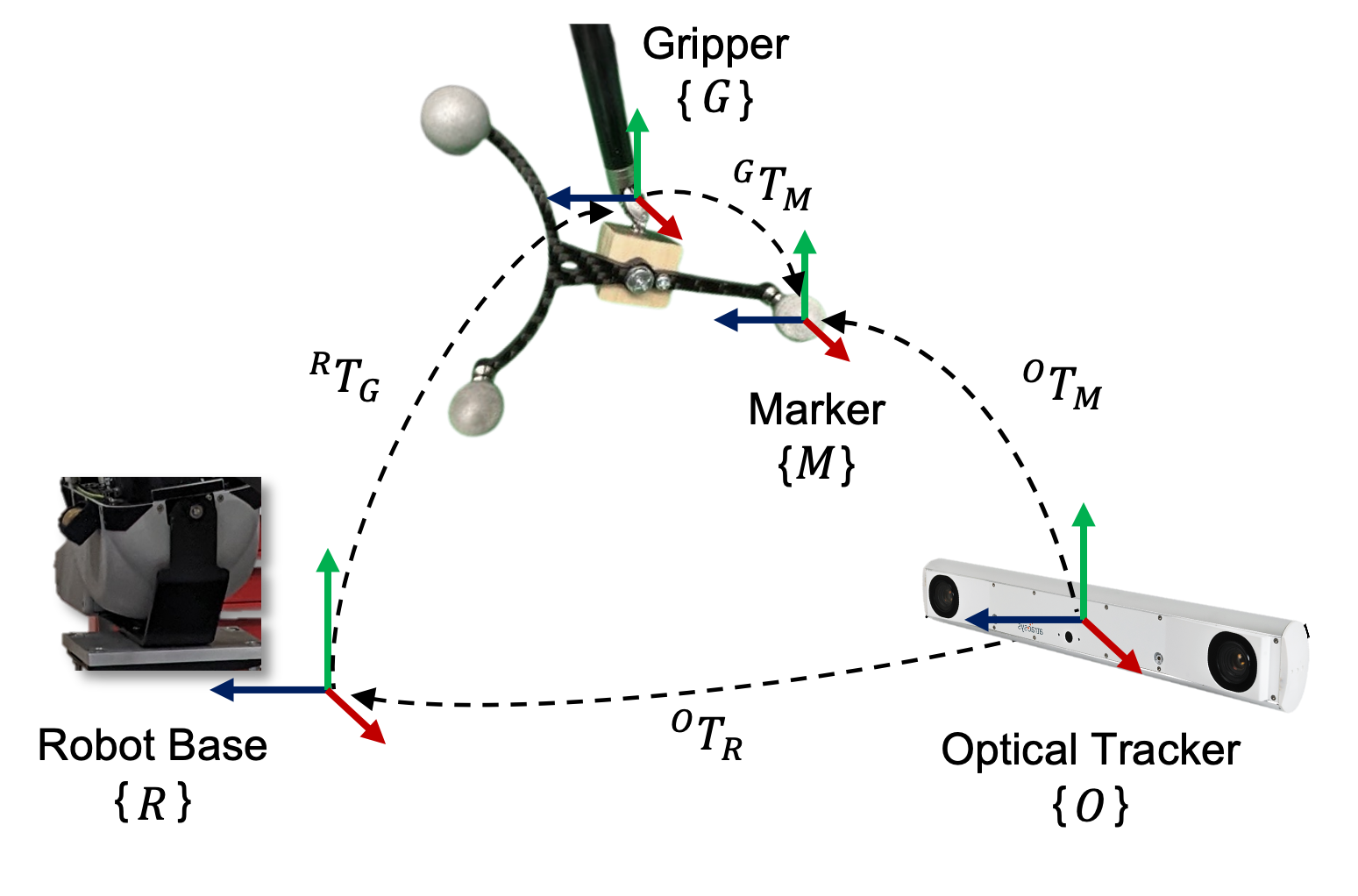}
    \caption{Transformation diagram that indicates how to calculate the robot's actual pose ($A_1$) with an optical tracker sensor.}
    \label{fig:transform_diagram}
\end{figure}

 As observed in Figure \ref{fig:transform_diagram}, there are two unknown transformations that need to be estimated before computing the actual pose of the robot using an external sensor: the transformation between the marker and the gripper control point ${}^{G}T_{M}$ and the transformation from the robot base to the tracker ${}^{O}T_{R}$. These transformations can be estimated by solving the hand-eye calibration problem in equation \ref{eq:hand_eye_calib}:

\begin{equation}
    {}^{R}T_{G} {}^{G}T_{M} = {}^{R}T_{O} {}^{O}T_{M}
   \label{eq:hand_eye_calib} 
\end{equation}

\noindent
where ${}^{R}T_{G}$ is the measured pose of the robot gripper with respect to the robot base and ${}^{O}T_{M}$ is a measurement of the location of the marker with respect to the optical tracker.  After performing the hand eye-calibration, the pose $A_1$ can be calculated using equation \ref{eq:A1comp}:

\begin{equation}
    A_1 = {}^{R}T_{O} {}^{O}T_{M} {}^{M}T_{G}
   \label{eq:A1comp} 
\end{equation}

\subsection{Neural network modeling} \label{metho-C}

To emulate realistic position errors on the simulated robot, two neural networks are trained to predict the error injection offsets. To train the neural networks, a dataset $S_1$, $M_1$, and $A_1$ is collected by moving the robot to multiple joint configurations. The first neural network $(NN_1)$ is trained to reproduce the error between $S_1$ and $M_1$, and therefore its training dataset follows equation \ref{eq:nn1_dataset}. $S_1^k$, $M_1^k$, and $A_1^k$ represent setpoint, measured, and actual pose at time $k$, respectively.

\begin{equation}
    \chi_{NN1} = \left\{\left(S_1^k,M_1^k-S_1^k\right)\mid k = [1 \ldots N]\right\},
    \label{eq:nn1_dataset}
\end{equation}

The second neural network $(NN_2)$ is trained to reproduce the error between $A_1$ and $M_1$, and therefore its training follows equation \ref{eq:nn2_dataset}. 

\begin{equation}
    \chi_{NN2} = \left\{\left(M_1^k,A_1^k-M_1^k\right)\mid k = [1 \ldots N]\right\},
    \label{eq:nn2_dataset}
\end{equation}

Given that the simulated and real robots are both controlled in joint space, the neural networks are designed to predict error injection offsets in joint space.  In this regard, both neural networks receive a feature vector of 6 inputs corresponding to a joint configuration and produce 6 outputs corresponding to the error injection offsets that need to be applied to each joint. The poses in the dataset are converted to joint configurations using the dVRK inverse kinematic function. Inputs and outputs are also normalized to have zero mean and unit standard deviation. Additionally, we experiment with augmenting the input feature vector with 6 more inputs corresponding to encoded previous measured positions. These additional features are added to account for the hysteresis-like nature of the non-linearities of cable-driven robots.

 \begin{figure}[ht]
    \centering
    \includegraphics[width=0.48\textwidth]{
    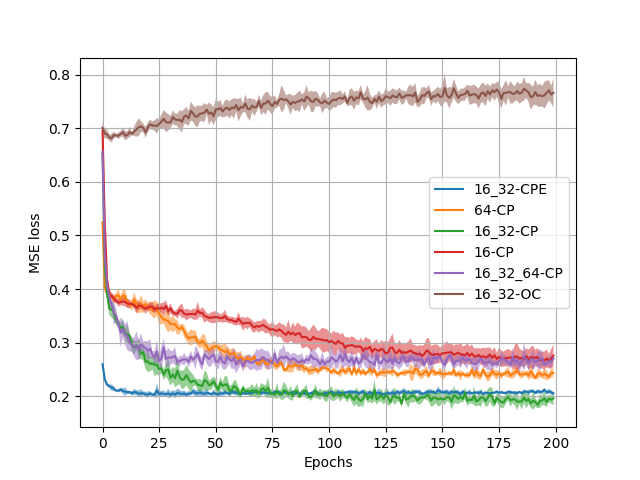}
    \caption{Hyperparameter optimization experiment for $NN_1$. The legend represents: \{Number of neurons in hidden layers\}-\{Type of input features\}. Three input features were considered:  Only current (OC), Current+previous (CP), and Current+previous encoded (CPE) (see subsection \ref{subsect-neural-net-modelling}). It is observed that adding previous measurements is necessary to reduce the validation loss.}
    \label{fig:hyper_parameter_optim}
\end{figure}

 \begin{figure*}[ht]
      \centering
      \includegraphics[width=1.0\textwidth]{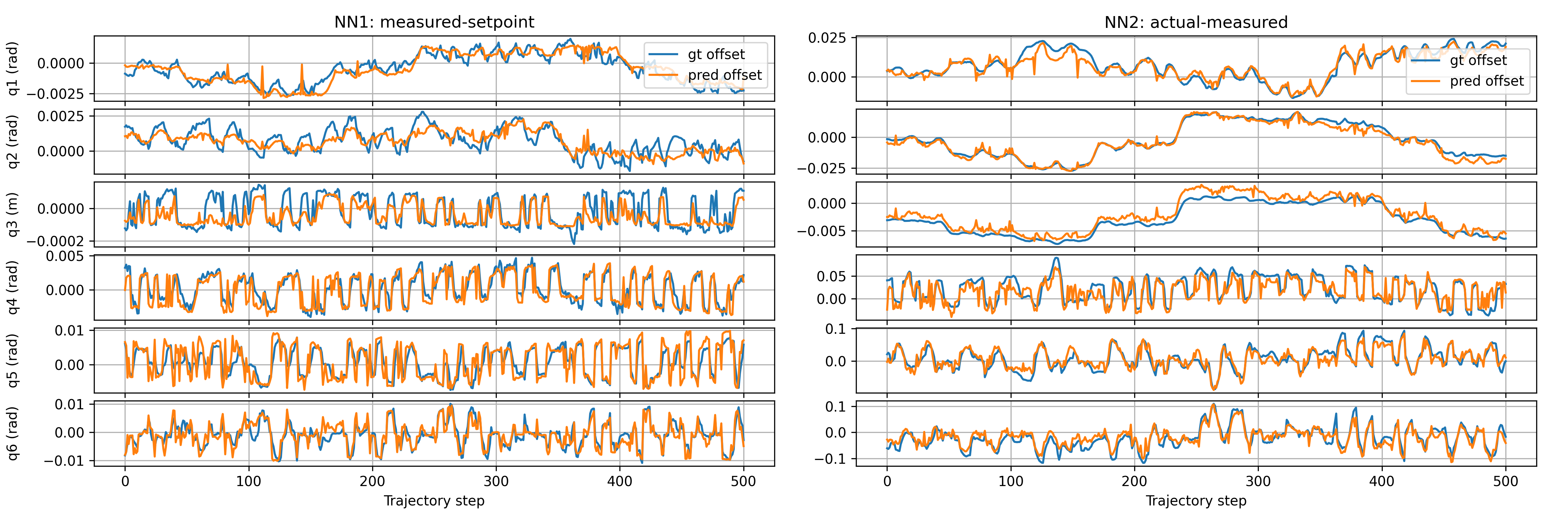}
      \captionof{figure}{Qualitative results for the neural networks. Left plot shows predictions for the controller error while the right plot shows predictions for kinematic/non-kinematic errors. Ground-truth offsets for each of the joints of the dVRK are shown in blue, while predicted offsets are shown in orange. Orange lines following closely the blue lines indicate that the neural networks are correctly modeling the robot errors. Notice that the dVRK has 5 rotational and 1 translational joints and therefore not all the offsets have the same units.}
      \label{fig:nn_predictions}
\end{figure*}
 
\subsection{Simulated robot motion error injection} \label{metho-D}

\ch{After training the two neural networks, realistic errors can be added to the simulated robot following a three-stage process.} In the first stage, the setpoint $(S_1)$ is given to the first neural network to predict the injection offset ($\alpha_1$), which indicates the error between $M_1$ and $S_1$. In the second stage, the joint configuration $S_1+\alpha_1$ is given to the second neural network to predict the injection offset ($\alpha_2$), which represents the error between $A_1$ and $M_1$. In the last stage, the new setpoint $S_1+\alpha_1 +\alpha_2$ is given to the simulated robot to initiate the motion. Assuming that the simulated robot does not have any control or kinematic errors, its final pose will be $A_2 = S_1+\alpha_1 +\alpha_2$ which should be approximately equal to $A_1$ (real robot's actual pose) (see Fig. \ref{fig:system_overview}). \ch{We note that although we are only testing the proposed error injection algorithm with a simulated dVRK, it is general enough to improve the realism of any simulated robot as long as it is possible to obtain training data from the real robot.}

\subsection{Evaluation metrics} \label{metho-E}

\ch{To evaluate differences between the simulated and real robot trajectories, translation (eq. \ref{eq:error_pos}) and rotation (eq. \ref{eq:error_rot})  error metrics are calculated.} The translation error ($E_T$) measures the Euclidean distance between two poses and can be calculated with 

\begin{equation} \label{eq:error_pos}
 E_{T} = ||\bar{t} - \hat{t}||   
\end{equation}

\noindent
where $\bar{t}$ and  $\hat{t}$ are the translational components of the real and simulated robot, respectively. The rotation error ($E_R$) measures the angle between two poses using the axis-angle representation of rotations \cite{Hodan2016OnEO} and can be calculated with 

\begin{equation} \label{eq:error_rot}
 E_{R} = \arccos\left(\frac{\mathrm{Tr} \left(\bar{R}\hat{R}^{-1} \right) - 1}{2}\right)   
\end{equation}

\noindent 
where  $\mathrm{Tr}()$ represents the trace operator and $\bar{R}$ and $\hat{R}$ represent the rotational components of the poses for the real and simulated robots, respectively.

\section{Experimental results}

\subsection{Experimental setup}
\begin{figure}[ht]
    \centering
    \includegraphics[width=0.48\textwidth]{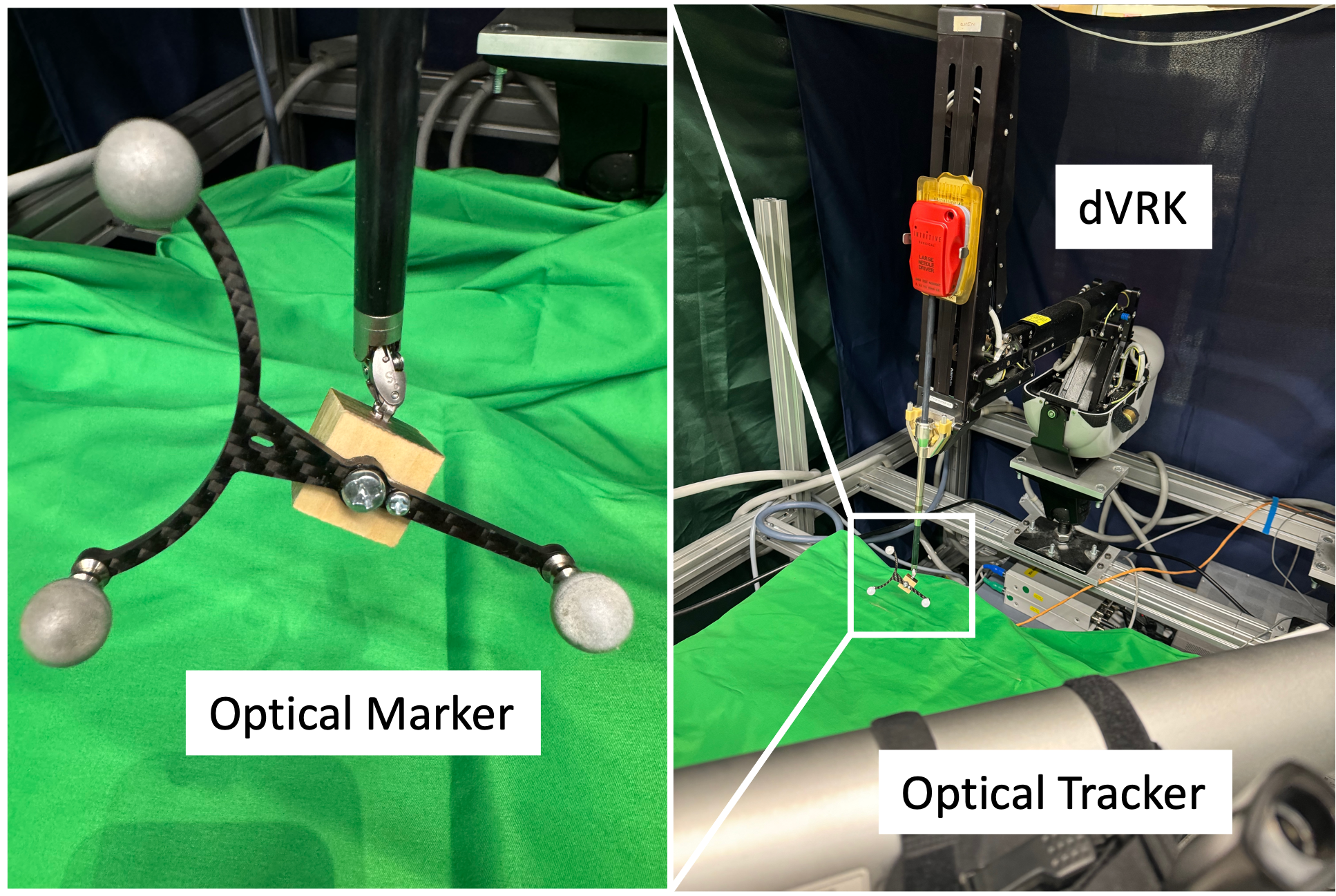}
    \caption{Experiment setup. The marker was rigidly attached to the tip of the tool, and it was tracked by the optical tracker.}
    \label{fig:experiment_setup}
\end{figure}

\ch{Our framework was tested on a dVRK which has two main components: a Patient Side Manipulator (PSM) with a large needle driver (LND) instrument and a Master Tool Manipulator (MTM) mainly used to teleoperate the PSM.} To collect data for the neural networks, an optically tracked marker was attached to the last link of the PSM as described in section \ref{metho-B}. To calculate the marker's pose, a fusionTrack 500 sensor (Atracsys, Switzerland) was mounted facing the robot approximately one meter away during the experiment (see Fig. \ref{fig:experiment_setup}). Regarding the simulation environment, we used the Asynchronous Multi-Body Framework (AMBF) \cite{MunawarAMBF} and we used the dVRK robot model that was developed in \cite{munawarOpen2022}. 

\subsection{Data collection and hand-eye calibration}

After setting up the equipment, the robot was programmed to execute a random trajectory of 3684 steps. The trajectory was generated by randomly sampling joint configurations as goal poses. The robot was commanded to take a step and then stop to take a measurement. To minimize excessive movements, intermediate steps were added between the beginning and random goal pose by linearly interpolating the two joint configurations. At every step of the motion, we recorded $S_1$ and $M_1$ from the robot and the location of the marker from the tracker. Utilizing this data, we performed a hand-eye calibration to calculate the robot's actual pose $(A_1)$. Hand-eye calibration was solved with the open-source solver in \cite{AyvaliPoseEStimation2024}. One additional trajectory was collected via teleoperation with the MTM for testing purposes.

\begin{table}[h]
\centering
\caption{Optimized hyperparameters used for the proposed Neural Network.}
\begin{tabular}{|c|c|c|c|}
\hline
batch size & learning rate & hidden layers & hidden neurons \\
\hline
32 & 0.0064 & 2 & 16, 32\\
\hline
\end{tabular}
\label{tab:hyperparam}
\end{table}

\subsection{Neural network modeling} \label{subsect-neural-net-modelling}

$NN_1$ and $NN_2$ are trained with the same data used to perform the hand-eye calibration and then tested with the teleoperated trajectory. The hand-eye calibration data is further split into a training dataset and a validation dataset to perform hyperparameter tuning. A simple multi-layer perception (MLP) with ReLU activation functions was used for both error regression tasks. While training the model, the Adam optimizer \cite{diederik_adamoptimization} and the MSE loss functions were used. Hyperparameter optimization experiments were performed to select the optimal network architecture, training parameters and input features. The optimized hyperparameters were training batch size, learning rate, number of hidden layers and number of hidden neurons in each layer. As for input features, three options were experimentally evaluated: 
 
\begin{enumerate}
\item \textbf{Only current (OC):} a 6-entry feature vector with only current measurements. 
\item \textbf{Current+previous (CP):} a 12-entry feature vector with current measurements and measurements from the previous step.
\item \textbf{Current+previous encoded (CPE):} a previous vector with current and previous measurements, but encoding the previous measurements as a vector of +1 or -1 indicating the direction of motion.   
\end{enumerate} 

 Figure  \ref{fig:hyper_parameter_optim} shows results from the hyperparameter experiments for $NN1$. From the plot, it can be seen that it is important to include information about the previous location to obtain a decreasing validation loss. Additionally, encoding the previous measurement sped up the convergence of the neural network. In terms of the network architecture, the most effective model had two hidden layers with 16 and 32 hidden neurons. The final set of hyperparameters can be seen in Table \ref{tab:hyperparam}. The same set of hyperparameters was used for both correction networks. Lastly, the optimized models are tested on the trajectory recorded via teleoperation. Figure \ref{fig:nn_predictions} shows qualitatively that the models are able to predict errors between $M_1$ and $S_1$ and $A_1$ and $M_1$. This illustrates that, although not a requirement, our error injection approach could potentially satisfy the stricter condition that $M_2^k \approx M_1^k$ and $A_2^k \approx A_1^k$ for all $k$.

\subsection{Correcting the motion of the virtual robot}

\begin{figure}[ht]
    \centering
    \includegraphics[width=0.48\textwidth]{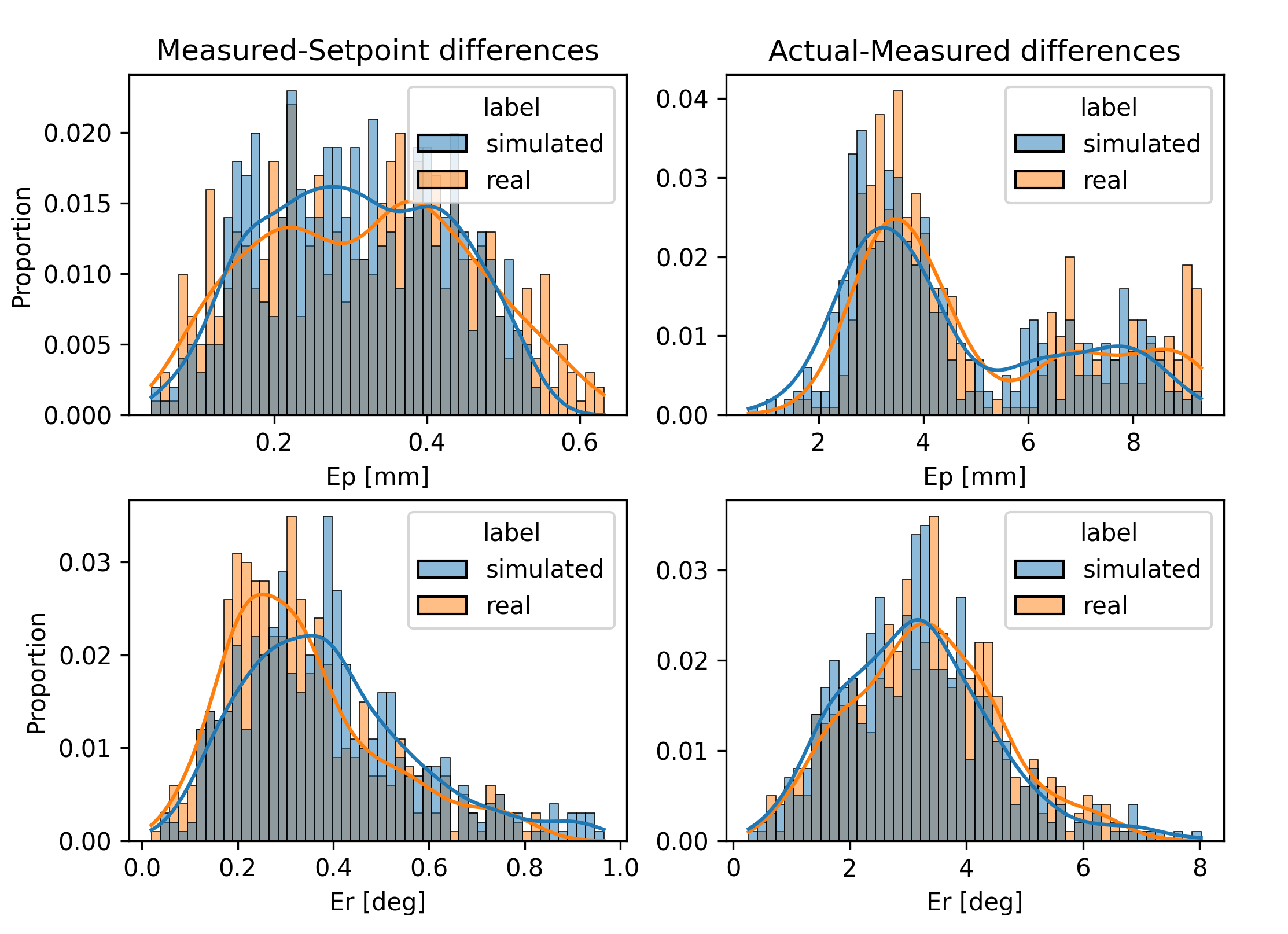}
    \caption{Error distributions of the real and simulated robots. It is observed that the simulated robot presents positioning errors that are very similar to the ones of the real robot after applying the corrections with the neural networks.}
    \label{fig:noise_patterns after corrections}
\end{figure}

Using the trained error models and setpoints from the teleoperated testing trajectory, $S_2$, $M_2$ and $A_2$ are generated for the simulated robot. To evaluate the performance of our system, we first compare the position error distributions of the simulated and real robots, i.e., the error distributions between setpoint and measured poses and measured and actual poses. As observed in Figure \ref{fig:noise_patterns after corrections}, the error distributions of the real and simulated robots are similar, indicating that the neural networks are effectively modeling the error from the real robot. Additionally, we compare the trajectories $A_1$ and $A_2$, with and without corrections from the neural network. As observed in Table \ref{tab:sim_real_differences}, the mean position error is reduced from $5.0\pm 2.0$ to $1.3 \pm 0.6$\,mm and the rotation error is reduced from $3.6 \pm 1.4$ to $1.7 \pm 0.7$ degrees. Both these results indicate that the neural networks allow the simulated robot to imitate the motion patterns of the real one. 

\begin{table}[h]
\centering
\caption{Comparison between $A_1$ and $A_2$ with and without correction. Lower values of error indicate that the actual trajectories of the real and simulated robots are more similar.}
\begin{tabular}{|ccccc|}
\hline
\multicolumn{5}{|c|}{\textbf{Differences between simulated and real actual poses}}                                                                                                                 \\ \hline
\multicolumn{1}{|c|}{\textbf{}}       & \multicolumn{2}{c|}{\textbf{Without corrections}}                                      & \multicolumn{2}{c|}{\textbf{With corrections}}                    \\ \hline
\multicolumn{1}{|c|}{}                & \multicolumn{1}{c|}{$\mathbf{E_T [mm]} $} & \multicolumn{1}{c|}{$\mathbf{E_R [deg]} $} & \multicolumn{1}{c|}{$\mathbf{E_T [mm]} $} & $\mathbf{E_R [deg]} $ \\ \hline
\multicolumn{1}{|c|}{\textbf{mean}}   & \multicolumn{1}{c|}{4.967}                & \multicolumn{1}{c|}{3.559}                 & \multicolumn{1}{c|}{1.316}                & 1.675                 \\ \hline
\multicolumn{1}{|c|}{\textbf{std}}    & \multicolumn{1}{c|}{2.048}                & \multicolumn{1}{c|}{1.416}                 & \multicolumn{1}{c|}{0.558}                & 0.719                 \\ \hline
\multicolumn{1}{|c|}{\textbf{median}} & \multicolumn{1}{c|}{4.023}                & \multicolumn{1}{c|}{3.510}                 & \multicolumn{1}{c|}{1.281}                & 1.616                 \\ \hline
\multicolumn{1}{|c|}{\textbf{max}}    & \multicolumn{1}{c|}{9.220}                & \multicolumn{1}{c|}{7.894}                 & \multicolumn{1}{c|}{5.734}                & 4.409                 \\ \hline
\end{tabular}
\label{tab:sim_real_differences}
\end{table}

\section{Summary and Future Work}

In this work, we propose a learning-based error approach to inject realistic noise patterns into a simulated robot. The proposed approach relies on neural networks to first predict different sources of error on a dVRK and then use these models to inject realistic error into the simulated robot. Ground-truth data to train the networks was generated with an optical tracker, which was used to calculate the actual position of the robot.  After training, the neural networks were successfully able to reduce position and orientation discrepancies between the simulated and real robots.

In future work, data collection will be extended to situations where the surgical robot is in contact with the environment. As described in \cite{chrysilla_compliance_2019}, applying external forces to surgical robotic instruments can lead to high pose errors due to mechanical compliance. In this regard, an additional neural network model could be trained to model the effects of external forces on the simulated robot. This will enable our approach to be more applicable to a broader range of surgical subtasks.


\section*{Acknowledgments and Disclosures}
This work was supported in part by NSF AccelNet award OISE-1927354.

\section*{Supplementary information} 
For more information, visit the project repository at \href{https://github.com/surgical-robotics-ai/ISMR2024-realistic-simulation-via-error-injection}{https://github.com/surgical-robotics-ai/ISMR2024-realistic-simulation-via-error-injection}

\bibliography{ISMR2024_dvrk_kin_calib}
\bibliographystyle{IEEEtran}

\end{document}